# Global Localization Based on 3D Planar Surface Segments Detected by a 3D Camera

Robert Cupec, Emmanuel Karlo Nyarko, Damir Filko,
Faculty of Electrical Engineering
J. J. Strossmayer University of Osijek
Osijek, Croatia
robert.cupec@etfos.hr

Andrej Kitanov, Ivan Petrović
Faculty of Electrical Engineering and Computing
University of Zagreb
Zagreb, Croatia
ivan.petrovic@fer.hr

*Abstract*—Global localization of a mobile robot using planar surface segments extracted from depth images is considered. The robot's environment is represented by a topological map consisting of local models, each representing a particular location modeled by a set of planar surface segments. The discussed localization approach segments a depth image acquired by a 3D camera into planar surface segments which are then matched to model surface segments. The robot pose is estimated by the Extended Kalman Filter using surface segment pairs as measurements. The reliability and accuracy of the considered approach are experimentally evaluated using a mobile robot equipped by a Microsoft Kinect sensor.

*Keywords—global localization; planar surfaces; Kinect; Extended Kalman Filter*

## I. Introduction

The ability of determining its location is vital to any mobile machine which is expected to execute tasks which include autonomous navigation in a particular environment. The basic robot localization problem can be defined as determining the robot pose relative to a reference coordinate system defined in its environment. This problem can be divided into two sub-tasks: initial global localization and local pose tracking. Global localization is the ability to determine the robot's pose in an a-priori or previously learned map, given no other information than that the robot is somewhere on the map. Local pose tracking, on the other hand, compensates small, incremental errors in a robot's odometry given the initial robot's pose thereby maintaining the robot localized over time. In this paper, global localization is considered.

There are two main classes of vision-based global localization approaches, appearance-based approaches and feature-based approaches.

In appearance-based approaches, each location in a robot's operating environment is represented by a camera image. Robot localization is performed by matching descriptors assigned to each of these images to the descriptor computed from the current camera image. The location corresponding to the image which is most similar to the currently acquired image according to a particular descriptor similarity measure is returned by the localization algorithm as the solution. The appearance-based techniques have recently been very intensively explored and some impressive results have been reported [1], [2].

In feature-based approaches, the environment is modeled by a set of 3D geometric features such as point clouds [3], points with assigned local descriptors [4], line segments [5], [6], surface patches [7] or planar surface segments [8], [9], [10], where all features have their pose relative to a local or a global coordinate system defined. Localization is performed by searching for a set of model features with similar properties and geometric arrangement to that of the set of features currently detected by the applied sensor. The robot pose is then obtained by registration of these two feature sets, i.e. by determining the rigid body transformation which maps one feature set onto the other.

An advantage of the methods based on registration of sets of geometric features over the appearance-based techniques is that they provide accurately estimated robot pose relative to its environment which can be directly used for visual odometry or by a SLAM system.

In this paper, we consider a feature-based approach which relies on an active 3D perception sensor. An advantage of using an active 3D sensor in comparison to the systems which rely on a 'standard' RGB camera is their lower sensitivity to changes in lighting conditions. A common approach for registration of 3D point clouds obtained by 3D cameras is *Iterative Closest Point* (ICP) [11], [12], [13]. Since this method requires a relatively accurate initial pose estimate, it can be used for local pose tracking and visual odometry, but it is not appropriate for global localization. Furthermore, ICP is not suitable for applications where significant changes in the scene are expected. Hence, we use a multi hypothesis Extended Kalman Filter (EKF).

Localization methods based on registration of 3D planar surface segments extracted from depth images obtained by a 3D sensor are proposed in [10] and [14]. In [14], a highly efficient method for registration of planar surface segments is proposed and its application for pose tracking is considered. In this paper, the approach proposed in [14] is adapted for global localization and its performance is analyzed. The environment model which is used for localization is a topological map consisting of local metric models. Each local model consists of planar surface segments represented in the local model reference frame. Such a map can be obtained by driving a robot with a camera mounted on it along a path the robot would follow while executing its regular tasks.

The rest of the paper is structured as follows. In Section II, the global localization problem is defined and a method for registration of planar surface segments is described which can be used for global localization. An experimental analysis of the proposed approach is given in Section III. Finally, the paper is concluded with Section IV.

## II. REGISTRATION OF PLANAR SURFACE SEGMENT SETS

The global localization problem considered in this paper can be formulated as follows. Given an environment map consisting of local models $M_1$, $M_2$, ..., $M_N$ representing particular locations in the considered environment together with spatial relations between them and a camera image acquired somewhere in this environment, the goal is to identify the camera pose at which the image is acquired. The term 'image' here denotes a depth image or a point cloud acquired by a 3D camera such as the Microsoft Kinect sensor. Let $S_{M,i}$ be the reference frame assigned to a local model $M_i$. The localization method described in this section returns the index $i$ of the local model $M_i$ representing the current robot location together with the pose of the camera reference frame $S_C$ relative to $S_{M,i}$. The camera pose can be represented by vector $w = \begin{bmatrix} \varphi^T & t^T \end{bmatrix}^T$, where $\varphi$ is a 3-component vector describing the orientation and $t$ is a 3-component vector describing the position of $S_C$ relative to $S_{M,i}$. Throughout the paper, symbol $R(\varphi)$ is used to denote the rotation matrix corresponding to the orientation vector $\varphi$.

The basic structure of the proposed approach is the standard feature-based localization scheme consisting of the following steps:

1. feature detection,
2. feature matching,
3. hypothesis generation,
4. selection of the best hypothesis.

Features used by the considered approach are planar surface segments obtained by segmentation of a depth image. These features are common in indoor scenes, thus making our approach particularly suited for this type of environments.

The surface registration algorithm considered in this paper is basically the same as the one proposed in [14]. The only difference is that instead of implementing visual odometry by registration between the currently acquired image and the previous image, global localization is achieved by registration between the currently acquired image and every local model $M_i$ in the map, where the initial pose estimate is set to the zero vector with a high uncertainty of the position and orientation. The proposed algorithm returns the pose hypothesis with the highest consensus measure [14] as the final result.

### A. Detection and Representation of Planar Surface Segments

Depth images acquired by a 3D camera are segmented into sets of 3D points representing approximately planar surface segments using a similar split-and-merge algorithm as in [15], which consists of an iterative Delaunay triangulation method followed by region merging. Instead of a region growing approach used in the merging stage of the algorithm proposed in [15], we applied a hierarchical approach proposed in [16] which produces less fragmented surfaces while keeping relevant details. By combining these two approaches a fast detection of dominant planar surfaces is achieved. The result is a segmentation of a depth image into connected sets of approximately coplanar 3D points each representing a segment of a surface in the scene captured by the camera. An example of image segmentation to planar surface segments is shown in Fig. 1.

The parameters of the plane supporting a surface segment are determined by least-square fitting of a plane to the supporting points of the segment. Each surface segment is assigned a reference frame with the origin in the centroid $t_F$ of the supporting point set and z-axis parallel to the supporting plane normal. The orientation of x and y-axis are defined by the eigenvectors of the covariance matrix $\Sigma_p$ computed from the positions of the supporting points of the considered surface segment within its supporting plane. The purpose of assigning reference frames to surface segments is to provide a framework for surface segment matching and EKF-based pose estimation explained in Section **Error! Reference source not**

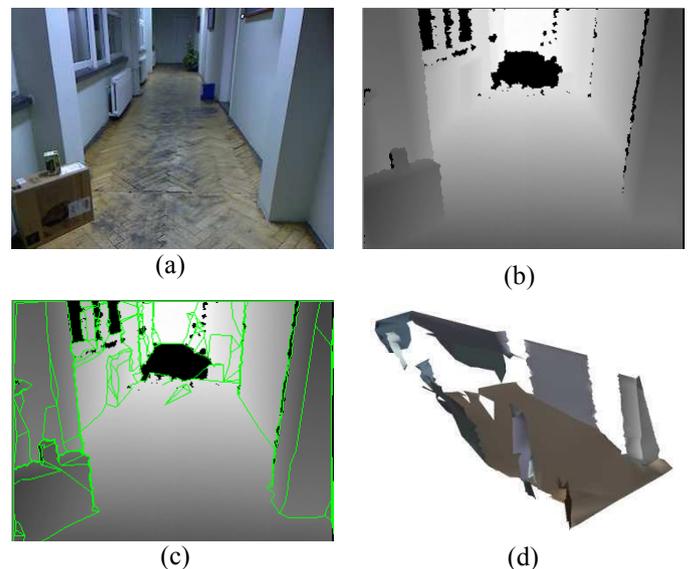

Fig. 1. An example of image segmentation to planar surface segments: (a) RGB image; (b) depth image obtained by Kinect, where darker pixels represent points closer to the camera, while black points represent points of undefined depth; (c) extracted planar surface segments delineated by green lines and (d) 3D model consisting of dominant planar surface segments.

**found.**.

Let the true plane be defined by the equation

$$^F\boldsymbol{n}^T \cdot {^F}\boldsymbol{p} = {^F}\rho, \qquad (1)$$

where $^F\boldsymbol{n}$ is the unit normal of the plane represented in the surface segment reference frame $S_F$, $^F\rho$ is the distance of the plane from the origin of $S_F$ and $^F\boldsymbol{p}$ is an arbitrary point represented in $S_F$. In an ideal case, where the measured plane is identical to the true plane, the true plane normal is identical to the z-axis of $S_F$, which means that $^F\boldsymbol{n} = [0, 0, 1]^T$, while $^F\rho = 0$. In a general case, however, the true plane normal deviates from the z-axis of $S_F$ and this deviation is described by the random variables $s_x$ and $s_y$, representing the deviation in directions of the x and y-axis of $S_F$ respectively, as illustrated in Fig. 2 for x direction. The unit normal vector of the true plane can then be written as

$$^F\boldsymbol{n} = \frac{1}{\sqrt{s_x^2 + s_y^2 + 1}} \begin{bmatrix} s_x & s_y & 1 \end{bmatrix}^T \qquad (2)$$

Furthermore, let the random variable $r$ represent the distance of the true plane from the origin of $S_F$, i. e.

$$^F\rho = r. \qquad (3)$$

The uncertainty of the supporting plane parameters can be described by the disturbance vector $\boldsymbol{q} = [s_x, s_y, r]^T$. We use a Gaussian uncertainty model, where the disturbance vector $\boldsymbol{q}$ is assumed to be normally distributed with 0 mean and covariance matrix $\Sigma_q$. Covariance matrix $\Sigma_q$ is a diagonal matrix with variances $\sigma_{sx}^2$, $\sigma_{sy}^2$ and $\sigma_r^2$ on its diagonal describing the uncertainties of the components $s_x$, $s_y$ and $r$ respectively. These variances are computed from the uncertainties of the supporting point positions, which are determined using a triangulation uncertainty model analogous to the one proposed in [17]. Let this uncertainty be represented by the 3×3 covariance matrix $\Sigma_C(\boldsymbol{p})$ assigned to each point position vector $\boldsymbol{p}$ obtained by a 3D camera. In order to achieve high computational efficiency, the centroid $\boldsymbol{t}_F$ of a surface segment is used as the representative supporting point and it is assumed that the uncertainties of all supporting points of this surface segment are similar to the centroid uncertainty. The variance $\sigma_r^2$ describing the uncertainty of the disturbance variable $r$ is computed as the uncertainty of the surface segment centroid $\boldsymbol{t}_F$ in the direction of the segment normal $\boldsymbol{n}$, computed by

$$\sigma_r^2 = \boldsymbol{n}^T \cdot \Sigma_C(\boldsymbol{t}_F) \cdot \boldsymbol{n}. \qquad (4)$$

The variances $\sigma_{sx}^2$ and $\sigma_{sy}^2$ describing the uncertainty of the segment plane normal are estimated using a simple model. The considered surface segment is represented by a flat 3D ellipsoid centered in the segment reference frame $S_F$, as illustrated in Fig. 3. Assuming that the orientation of the surface segment is computed from four points at the ellipse perimeter which lie on the axes x and y of $S_F$, as illustrated in Fig. 3, the uncertainty of the surface segment normal can be computed from the position uncertainties of these four points. According to this model the variances $\sigma_{sx}^2$ and $\sigma_{sy}^2$ can be computed by

$$\sigma_{sx}^2 \approx \frac{\sigma_r^2}{\lambda_1 + \sigma_r^2}, \qquad \sigma_{sy}^2 \approx \frac{\sigma_r^2}{\lambda_2 + \sigma_r^2}, \qquad (5)$$

where $\lambda_1$ and $\lambda_2$ are the two largest eigenvalues of the covariance matrix $\Sigma_p$. Alternatively, a more elaborate uncertainty model can be used, such as those proposed in [6] and [10].

Finally, a scene surface segment is denoted in the following by the symbol $F$ associated with the quadruplet

$$F = \left( {^C}\boldsymbol{R}_F, {^C}\boldsymbol{t}_F, \Sigma_q, \Sigma_p \right), \qquad (6)$$

where $^C\boldsymbol{R}_F$ and $^C\boldsymbol{t}_F$ are respectively the rotation matrix and translation vector defining the pose of $S_F$ relative to the camera coordinate system $S_C$. Analogously, a local model surface segment is represented by

$$F' = \left( {^M}\boldsymbol{R}_{F'}, {^M}\boldsymbol{t}_{F'}, \Sigma_{q'}, \Sigma_{p'} \right), \qquad (7)$$

where index $M$ denotes the local model reference frame.

*B. Initial Feature Matching*

The pose estimation process starts by forming a set of surface segment pairs $(F, F')$, where $F$ is a planar surface segment detected in the currently acquired image and *F'* is a local model surface segment. A pair (*F*, *F'*) represents a correct correspondence if both *F* and *F'* represent the same surface in the robot's environment. The surface segments detected in the currently acquired image are transformed into the local model reference frame using an initial estimate of the camera pose relative to this frame and every possible pair (*F*, *F'*) of surface segments is evaluated according to the coplanarity and overlap criteria explained in [14]. These two criteria take into account the uncertainty of the assumed camera pose. In the experiments reported in this paper, the initial robot pose estimate used for feature matching is set to zero vector with the uncertainty described by a diagonal covariance matrix $\Sigma_{w,match} = \text{diag}([\sigma_\phi^2, \sigma_t^2, \sigma_t^2])$, where $\sigma_\alpha = 20°$ and $\sigma_t = 1$ m describe the uncertainty of the robot orientation and position in xy-plane of the world reference frame respectively. This uncertainty is propagated to the camera reference frame, taking into account the uncertainty of the camera inclination due to the uneven floor. The deviation of the floor surface from a perfectly flat surface is modeled by zero-mean Gaussian noise with standard deviation $\sigma_f = 5$ mm.

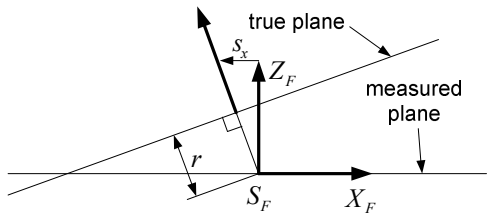

Fig. 2. Displacement of the true plane from the measured plane.

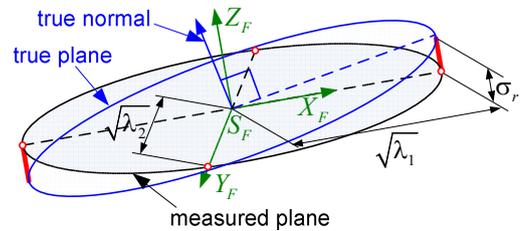

Fig. 3. Plane uncertainty model.

*C. Hypothesis Generation*

Given a sequence of correct pairs (*F*, *F'*), the camera pose relative to a local model reference frame can be computed using the EKF approach. Starting from the initial pose estimate, the pose information is corrected using the information provided by each pair (*F*, *F'*) in the sequence. After each correction step, the pose uncertainty is reduced. This general approach is applied e.g. in [18] and [5]. Some specific details related to our implementation are given in the following.

Let (*F*, *F'*) be a pair of corresponding planar surface segments. Given a vector $^{F'}p$ representing the position of a point relative to $S_{F'}$, the same point is represented in $S_F$ by

$$^{F}p = {^{C}R_F^T}\left({R^T(\phi)}\left({^{M}R_{F'}}\,{^{F'}p} + {^{M}t_{F'}} - t\right) - {^{C}t_F}\right), \quad (8)$$

where $w = \begin{bmatrix} \phi^T & | & t^T \end{bmatrix}^T$ is an estimated camera pose relative to a local model reference frame. By substituting (8) into (1) we obtain

$$^{F'}n^T \cdot {^{F'}p} = {^{F'}\rho} \quad (9)$$

where

$$^{F'}n = {^{M}R_{F'}^T}\,R(\phi)\,{^{C}R_F}\,{^{F}n}, \quad (10)$$

$$^{F'}\rho = {^{F}\rho} + {^{F}n^T} \cdot {^{C}R_F^T}\left({^{C}t_F} + R^T(\phi)\left(t - {^{M}t_{F'}}\right)\right). \quad (11)$$

Vector $^{F'}n$ and value $^{F'}\rho$ are the normal of *F* represented in $S_{F'}$ and the distance of the plane supporting *F* from the origin of $S_{F'}$ respectively. The deviation of the plane supporting the scene surface segment from the plane containing the local model surface segment can be described by the difference between the plane normals and their distances from the origin of $S_{F'}$. Assuming that *F* and *F'* represent the same surface, the following equations hold

$$^{F'}n = {^{F'}n'}, \quad (12)$$

$$^{F'}\rho = {^{F'}\rho'}, \quad (13)$$

where $^{F'}n'$ and $^{F'}\rho'$ are the parameters of the plane supporting *F'* represented in reference frame $S_{F'}$. Since $^{F'}n$ and $^{F'}n'$ are unit vectors with two degrees of freedom, it is appropriate to compare only their two components. We choose the first two components to formulate the coplanarity constraint

$$\begin{bmatrix} \begin{bmatrix} 1 & 0 & 0 \\ 0 & 1 & 0 \end{bmatrix}\left({^{F'}n} - {^{F'}n'}\right) \\ \hdashline {^{F'}\rho} - {^{F'}\rho'} \end{bmatrix} = \mathbf{0} \quad (14)$$

Note that the vector on the left side of equation (14) is actually a function of the disturbance vectors *q* and *q'* representing the uncertainty of the parameters of the planes supporting *F* and *F'* respectively, the pose *w* and the estimated poses of *F* and *F'* relative to $S_C$ and $S_M$ respectively. Hence, (14) can be written as

$$h(F, F', w, q, q') = \mathbf{0}. \quad (15)$$

Equation (15) represents the measurement equation from which EKF pose update equations can be formulated using the general approach described in [18].

This EKF-based procedure will give a correct pose estimate assuming that the sequence of surface segment pairs used in the procedure represent correct correspondences. Since the initial correspondence set usually contains many false pairs, a number of pose hypotheses are generated from different pair sequences and the most probable one is selected as the final solution. We use the efficient hypothesis generation method described in [14]. The result of the described pose estimation procedure is a set of pose hypotheses ranked according to a consensus measure explained in [14]. Each hypothesis consists of the index of a local model to which the current depth image is matched and the camera pose relative to the reference frame of this model.

III. EXPERIMENTAL EVALUATION

In this section, the results of an experimental evaluation of the proposed approach are reported. We implemented our system in C++ programming language using OpenCV library [19] and executed it on a 3.40 GHz Intel Pentium 4 Dual Core CPU with 2GB of RAM. The algorithm is experimentally evaluated using 3D data provided by a Microsoft Kinect sensor mounted on a wheeled mobile robot Pioneer 3DX also equipped with a laser range finder SICK LMS-200. For the purpose of this experiment, two datasets were generated by manually driving the mobile robot on two different occasions through a section of a previously mapped indoor environment of the Department of Control and Computer Engineering, Faculty of Electrical Engineering and Computing (FER), University of Zagreb. The original depth images of resolution 640 × 480 were subsampled to 320 × 240. Fig. 4 shows the previously mapped indoor environment generated with the aid of SLAM using data from the laser range finder together with the trajectory of the robot when generating the first dataset.

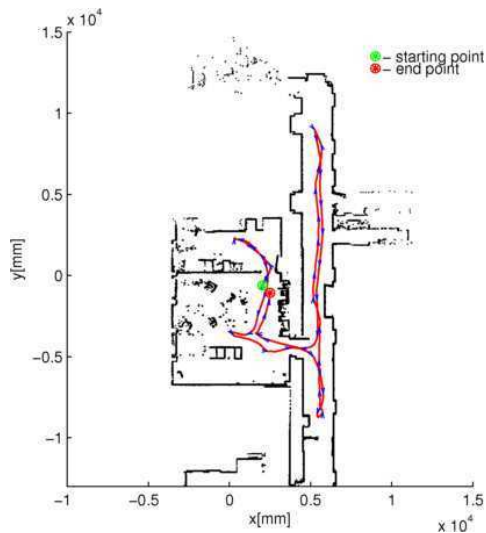

Fig. 4. The map of the Department of Control and Computer Engineering, (FER, Zagreb) obtained using SLAM and data from a laser range finder and the trajectory of the wheeled mobile robot while generating images used in creating the topological map.

The first dataset consists of 444 RGB-D images recorded along with the odometry data. The corresponding *ground truth* data as to the *exact* pose of the robot in the global coordinate frame of the map was determined using laser data and Monte Carlo localization. These images were used to create the environment model – a database of local metric models with topological links. This environment model or topological map consisted of 142 local models, generated such that the local model of the first image was automatically added to the map and every consecutive image or local model added to the map satisfied at least one of the following conditions: (1) the translational distance between the candidate image and the latest added local model in the map was at least 0.5 m or (2) the difference in orientation between the candidate image and the latest added local model in the map was at least 15°.

The trajectory of the robot during the generation of the second sequence was not the same as the first sequence but covered approximately the same area. With the aid of odometry information from the robot encoders, the second sequence was generated by recording RGB-D images every 0.5 m or 5° difference in orientation between consecutive images. The corresponding ground truth data was determined using laser data and Monte Carlo localization and recorded as well. This second dataset consisted of a total of 348 images.

The global localization procedure was performed for all the images in both datasets with the topological map serving as the environment model. Thus, all 792 images were tested, among which 142 were used for model building. As explained in Section II, each generated hypothesis provides the index of the local model from the topological map as well as the relative robot pose corresponding to the test image with respect to the local model. This robot pose is referred to herein as a *calculated* pose. By comparing the calculated pose of the test image to the corresponding ground truth data, the accuracy of the proposed approach can be determined. For each test image, a *correct* hypothesis is considered to be one where: (1) the translational distance between the calculated pose and the ground truth data is at most 0.2m and; (2) the absolute difference in orientation between the calculated pose and the ground truth is at most 2°. Using this criterion, the effectiveness of the proposed global localization method can be assessed not only on the basis of the accuracy of the solutions, but also on the minimum number of required hypotheses that need to be generated in order to obtain at least one correct hypothesis. Examples of images from both datasets are given in Fig. 5.

An overview of the results of the initial global localization experiment is given in Table I. Of the 792 test images, the proposed approach was not able to generate any hypothesis in 30 cases. In all 30 cases, the scenes were deficient in information needed to estimate all six degrees of freedom (DoF) of the robot's pose. Such situations normally arise when the camera is directed towards a wall or door at a close distance, e.g. while the robot is turning around in a corridor.

TABLE I. GLOBAL LOCALIZATION RESULTS

|  | Number of images | Percentage (%) |
|---|---|---|
| Total | 792 | 100.00 |
| No hypothesis | 30 | 3.79 |
| No correct hypothesis | 44 | 5.56 |
| Correct hypothesis | 718 | 90.65 |

In 44 cases, no correct hypothesis was generated by the proposed approach. There were two main reasons for such a situation: (1) the topological map did not contain a local model covering the scene of the test image; (2) the existence of repetitive structures in the indoor environment. An example of such situations is a pair of scenes shown in the last column

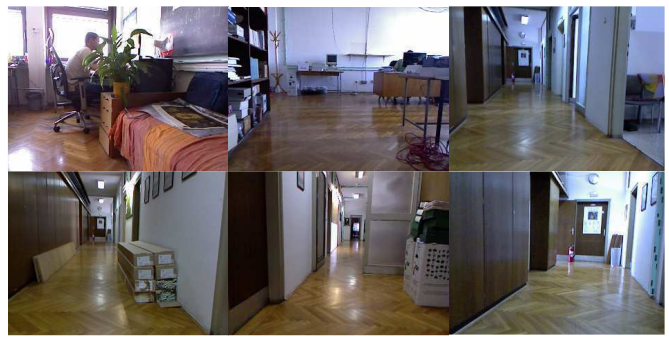

Fig. 5. Examples of images used in the global localization experiment.

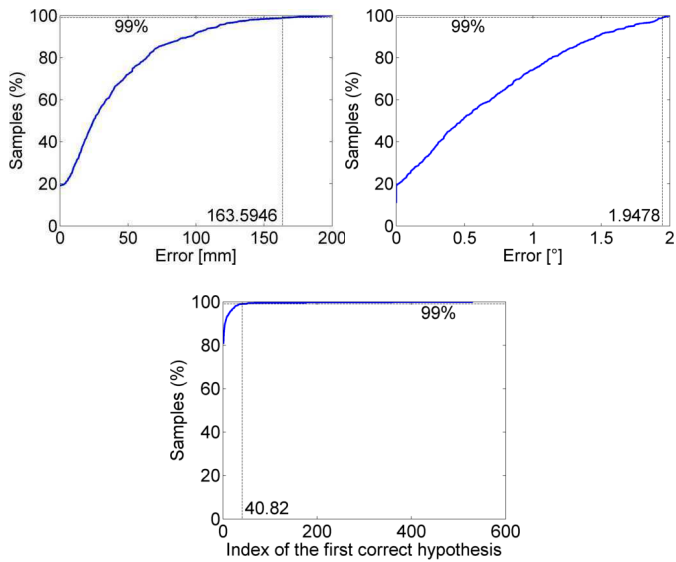

Fig. 6    Normalized cumulative histogram of the error in position (top-left) error in orientation (top-right) index of the first correct hypothesis (bottom).

of Fig. 5, where one can notice the similar *repeating* doorways on the left side of the corridor.

The accuracy of the proposed approach is determined using the 718 images with a correct hypothesis. The results are shown statistically in Table II as well as in Fig. 6, in terms of the absolute error in position and orientation between the correct pose and corresponding ground truth pose of the test sequence images as well as the index of the first correct hypothesis. The error bounds as well as the number of highest ranked hypotheses containing at least one correct hypothesis for 99% of samples are specially denoted in Fig. 6.

TABLE II.    STATISTICAL DETAILS OF THE GLOBAL LOCALIZATION POSE ERROR AND THE INDEX OF THE FIRST CORRECT HYPOTHESIS

|      | Translation Error (mm) | Orientation Error (°) | Index of the first correct hypothesis |
|------|-----------------------:|----------------------:|--------------------------------------:|
| Avg. | 36.83                  | 0.62                  | 4.29                                  |
| Std. | 38.22                  | 0.56                  | 25.53                                 |
| Max. | 199.65                 | 1.99                  | 530.00                                |

## IV. CONCLUSION

In this paper a global localization approach based on the environment model consisting of planar surface segments is discussed. Planar surface segments detected in the local environment of the robot are matched to the planar surface segments in the model and the robot pose is estimated by registration of planar surface segment sets based on EKF. The result is a list of hypotheses ranked according to a measure of their plausibility. The considered approach is experimentally evaluated using depth image sequences acquired by a Microsoft Kinect sensor mounted on a mobile robot. The analyzed approach generated at least one correct pose hypothesis in 90% of cases. On average, the first correct hypothesis is the 4$^{th}$ ranked hypothesis. For the highest ranked correct hypotheses, the error in position was on average approximately 37 mm, while the difference in orientation was on average approximately 0.6°. For 99% of these hypotheses, the pose error was at most 164 mm and 1.9°.